\documentclass{article}
\usepackage{graphicx} 
\usepackage{caption}
\usepackage[margin=1in]{geometry}
\usepackage{authblk}
\usepackage{setspace}
\usepackage{amsmath}
\usepackage{indentfirst}
\usepackage{amsfonts} 
\usepackage{float}
\usepackage{tabularx} 
\usepackage{booktabs} 
\usepackage{array} 
\usepackage{csquotes}
\usepackage{hyperref}
\usepackage{subcaption}
\hypersetup{
    colorlinks = true, 
    linkcolor = black, 
    citecolor = blue, 
    urlcolor = blue 
}
\usepackage[style=apa, backend=biber, alldates=year, doi=true, uniquename=false, uniquelist=false]{biblatex} 

\usepackage{longtable}
\usepackage{verbatim}

\title{What if? Causal Machine Learning in Supply Chain Risk Management}
\author[a]{Mateusz Wyrembek\thanks{Corresponding author email: mateusz.wyrembek@phd.ue.poznan.pl}}
\author[b]{George Baryannis}
\author[c,d]{Alexandra Brintrup}
\affil[a]{Department of Logistics, Poznań University of Economics and Business, PL}
\affil[b]{Department of Computer Science, University of Huddersfield, UK}
\affil[c]{Supply Chain AI Lab, University of Cambridge, UK}
\affil[d]{Data Centric Engineering, Alan Turing Institute, UK}
\date{}

\begin{document}

\maketitle

\noindent\makebox[\linewidth]{\rule{\textwidth}{1pt}} 

\begin{center}
\textbf{ABSTRACT}
\end{center}

\noindent The penultimate goal for developing machine learning models in supply chain management is to make optimal interventions. However, most machine learning models identify correlations in data rather than inferring causation, making it difficult to systematically plan for better outcomes. In this article, we propose and evaluate the use of causal machine learning  for developing supply chain risk intervention models, and demonstrate its use with a case study in supply chain risk management in the maritime engineering sector. Our findings highlight that causal machine learning  enhances decision-making processes by identifying changes that can be achieved under different 
supply chain interventions, allowing "what-if" scenario planning. We therefore propose different machine learning developmental pathways for for predicting risk,  and planning for interventions to minimise risk and outline key steps for supply chain researchers to explore causal machine learning.

\vspace{5pt}

\noindent \textbf{\textit{Keywords:}} Supply Chain Risk Management, Supply Chain, Machine Learning, Causal Inference, Causal Machine Learning, Delay Prediction

\noindent\makebox[\linewidth]{\rule{\textwidth}{1pt}}

\setlength{\parindent}{15pt} 

\section{Introduction}

Advances in artificial intelligence (AI) and machine learning (ML) have opened novel paths to inferring new knowledge from supply chain data, yielding advances in classical problems such as demand forecasting \parencite{Soori2023}, and entirely new areas such as delay prediction and uncovering hidden supply chain dependencies \parencite{Kosasih_2021, Wyrembek_2023}.

While these advances have opened up new opportunities, managers often highlight the "black-box" problem when using AI to make decisions as they are ultimately held responsible if anything goes wrong. The lack of transparency and interpretability in ML models, as well as the provision of comprehensive explanations of AI outputs can lead to a lack of trust and reluctance to fully integrate these technologies into decision-making processes.  For instance, consider a model  built for predicting a supplier delivery delay. A deep learning model, although providing insights into patterns and trends, would fall short in identifying the underlying causal mechanisms that drive a delay, leading to potential oversights in risk mitigation strategies. For a planner, the ultimate goal of developing an ML model might be to make an intervention such that the foreseen delay is minimised. There might be a number of alternative interventions. For example, one might order from an alternative supplier, order a different quantity or order at a different time period. The planner thus would need to estimate the effect of one of these interventions on the outcome. Doing so, would require a shift from identifying correlations in data, to identifying the underlying causal mechanisms that drive an outcome.

Our aim in this article is not to criticise the use of AI and ML for the development of supply chain risk prediction models, but to highlight that prediction and intervention models have different developmental paths and intended uses. In domains such as healthcare, this observation has motivated research in causal machine learning (CML), which combines machine learning with causal inference to estimate intervention effects and optimise the allocation of interventions amongst groups of intervention recipients. In tandem, there have been calls for more explainable AI in supply chain management \parencite{kosasih2024review}, and inquiries on how best to use ML models for making optimal interventions. These calls emphasise the growing need for methodologies that not only predict outcomes but also provide actionable insights that can inform strategic decisions. However, to date there have been no studies on CML in supply chain management - a methodology which might help address both issues. 

CML is used to estimate a causal quantity of change in a given outcome due to the result of an intervention \parencite{kaddour2022causal}. A key benefit of CML is that it allows for the estimation of individualised interventions on outcome variables, so that decision-making can be tailored to a given context. CML can handle high dimensional unstructured data and establish which sub groups of data are best recipients of a given intervention. As such, one can perform "what if" reasoning to evaluate how outcomes will change due to an intervention. 

 In this article we outline the key steps in developing CML for supply chain intervention models, using a  case study in supply chain risk management (SCRM) in the maritime engineering sector. Our hope is that doing so presents an illustrative case that can motivate the use of CML in SCRM tasks where optimal interventions are required. We then discuss the limitations of CML, propose directions for future research and provide recommendations for the reliable use of CML.

\section{Causal Machine Learning}

Causal machine learning is different from traditional predictive ML, in that traditional ML aims at predicting outcomes, while CML quantifies changes in outcomes due to a given intervention (termed as "treatment" in CML). 

Although methods for estimating intervention effects have a long tradition in the statistical literature,  CML offers distinct benefits by combining ML with causal inference. For example, methods from classical statistics often assume prior knowledge about the association between input and output variables, such as linear dependencies. Such knowledge is often not available or assumptions are unrealistic, especially for high-dimensional datasets. 

Estimating intervention effects from data is non trivial due to a number of challenges. The first of these is the causal inference problem, where we aim to infer causality from data. Consider the following case of developing an ML model to predict quality defects in a product. In the developed model, it appears that the use of a new supplier is correlated to an increase in defects. If we look closely, we could see that there has been a change in the design of a component of the product, leading to defects. The design change, here, is a \textit{confounding variable}, which is related to both the input variable (supplier) and the output variable (defects) in a model and can distort the causal relationship between them, giving rise to an apparent causal relationship that is actually spurious. A further complication occurs when intervention effects for individual problem cases may not be observable  because one can only observe the factual outcome under a given intervention (e.g. a defect after a supplier has been chosen), but one cannot observe the counterfactual outcome under a different intervention scenario (what the outcome would be, had a different supplier been chosen). 


CML seeks to address these limitations by a \textit{double/debiased machine learning (DML)} framework. DML was first introduced by \textcite{chernozhukov2018double} and is based on the principles of the Frisch-Waugh-Lovell Theorem \parencite{frisch1933partial, lovell1963seasonal}. This theorem states that, given the linear model Y=\(\beta_0\)+
\(\beta_1\)D+\(\beta_2\)Z+U, the two following approaches for estimating \(\beta_1\) yield the same result:
\begin{itemize}
    \item Linear regression of Y on D and Z, using Ordinalry Least Squares (OLS).
    \item Three-step procedure: 1) regress D on Z; 2) regress Y on Z; 3) regress the residuals from step 2 on the residuals from step 1 for getting an estimate \(\beta_1\) (all regressions using OLS).
\end{itemize}

In a similar manner, in DML we can proceed as follows:
\begin{enumerate}
    
 \item Predict D based on Z using a suitable ML algorithm;
 \item Predict Y based on Z using a suitable ML algorithm;
 \item Linear regression of the residuals from step 2 on the residuals from step 1, for getting an estimate of \(\theta_0\).
\end{enumerate}

\noindent A wide range of ML approaches, such as random forests, boosting or neural networks, can be utilised for the prediction tasks above \parencite{Bach_2022_JMLR}.

This procedure ensures that the model from step 3 is ``orthogonalised'', yielding an unbiased, root \textit{n}-consistent estimator that is free of confounders. In essence, regression on the residuals yields the effect of confounders (called \textit{nuisance functions}), which are not of primary interest to the modeller. This is also called the Neyman orthogonality condition. To estimate the causal parameter \(\theta_0\), the empirical analog of the moment condition \(E(\psi(W;\theta_0,\eta_0)) = 0\) is solved, where \(W = (Y,D,X)\), \(\psi(\cdot)\) represents the score function, and \(\eta_0 = (g_0,m_0)\) are known as the nuisance functions. The score function meets the Neyman orthogonality condition if \(\partial_\eta E(\psi(W;\theta_0,\eta))|_{\eta=\eta_0} = 0\), with \(\partial_\eta\) indicating the pathwise Gateaux derivative operator. 

As discussed earlier, the use of ML estimators to approximate the nuisance functions \(\eta_0\) in steps 2 and 3 introduces a regularisation bias. Imposing the Neyman orthogonality condition to our score function (and thus to our estimators of \(\theta_0\) as well as \(\eta_0\) makes our estimator of \(\theta_0\) free of the regularisation bias.

A final ingredient of DML is the strategy of sample splitting. Estimating the nuisance functions on the same dataset used for the parameter \(\theta_0\) can lead to overfitting bias. However, this issue can be mitigated through sample splitting, meaning that the nuisance functions \(\eta_0\) are estimated using one part of the dataset, while the score function for the desired parameter \(\theta_0\) is computed with the remaining data. Implementing sample splitting across \(K\) folds and adopting repeated cross-fitting enhances the robustness of the estimates \parencite{bach2024doubleml}. For this, we follow the procedure below:
\begin{enumerate}
   
\item randomly partition our data into two subsets
    \item fit our ML models for D and Y on the first subset.
 \item estimate \(\theta_0,_1\) in the second subset using the models obtained in step 2.
 \item fit our ML models in the second subset.
 \item estimate \(\theta_0,_2\) in the first subset, using the models obtained in step 4.
 \item obtain our final estimator \(\theta_0\) as an average of \(\theta_0,_1\) and \(\theta_0,_2\)
\end{enumerate}

The choice of the ML model itself is case dependent. In the case study presented in the next section, we choose an interactive regression model (IRM). We do so because our treatment (intervention variable) is binary which an IRM can handle in the DML framework. We will discuss the steps involved in applying DML, including IRM, within the context of the case study next.

\section{Causal Machine Learning applied to Supply Chain Risk Management: An illustrative case study }

\subsection{Related research}

Leaning on the observation that most of extant SCRM research focuses on post-delay mitigation strategies, recent advances in ML research have presented a promising avenue for predicting risk. Based on this, one of the earliest applications of ML in supply chain management was on supplier delay prediction \parencite{brintrup2020supply}, where Material Requirements Planning (MRP) data was used to predict order delays. Several follow on studies investigated classification models  (e.g. \textcite{LOLA, Sarbas2023, WANI2022731, douaioui2024enhancing, BASSIOUNI2024}), and regression based approaches (e.g. \textcite{STEINBERG2023100003, Gabellini_2024, balster2020eta, Shah_23}) to predict delay severity and lead times.  Other research focused on the role of information sharing for delay prediction among multiple buyers \parencite{Zheng2023}, efficacy of ML to predict delays during Covid-19 (e.g.  \textcite{Zaghdoudi_2022, mariappan2022large, mariappan2023using}). Delay prediction since then has been applied to aerospace, maritime, eCommerce, and pharmaceutical case studies.

Whilst delay prediction has been a popular area of research, authors raised the lack of interpretability on the use of ML in evaluating risk, which prevent these models to be used in practical contexts \parencite{baryannis2019predicting, brintrup2023trustworthy}. Authors proposed the use of interpretable ML models such as decision trees \parencite{baryannis2019predicting}, multi-criteria decision-making \parencite{WyrembekBaryannis2024,ABDULLA2023100342,ABDULLA2024100074} and neurosymbolic methods \parencite{kosasih2024review}. However, they also pointed out the trade off between performance and interpretability - black box ML  models were often more accurate but their findings were not easily interpretable. 

In this study we take an alternative view on this conundrum, highlighting the different developmental paths for the use of ML in SCRM. The first of these is prediction, which is to use past data to make estimations on the future, where the goal of the predictive process is to make as accurate estimations as possible. The second developmental path, which is less well studied, is causal inference. In domains such as healthcare analytics causal inference has been used for interpreting causes of change in an outcome variable of interest, and using this interpretation for planning case-specific interventions, which researchers in SCRM have not yet made full use of. 

Although ML models hitherto proposed for SCRM can accurately predict whether a supply chain delay will occur and how long it might last, they are not effective in determining causes of delays, which is a prerequisite in mitigation planning. Current models that predict risk assume the onus of finding causality on the practitioner. As causality is unknown or at least subjectively attributed, once a delay is predicted, the practitioner can only plan for reactive steps that occur after the delay, rather than focus on preventing it through mitigation.  

To address this issue, we propose complementing ML based delay prediction with causal inference, such that effective control can be implemented prior to the risk being manifested, when possible. We use the following problem setting for exploring this complementary avenue.  

\subsection{Problem setting}

\begin{figure}[H]
    \centering
    \includegraphics[width=0.7\linewidth]{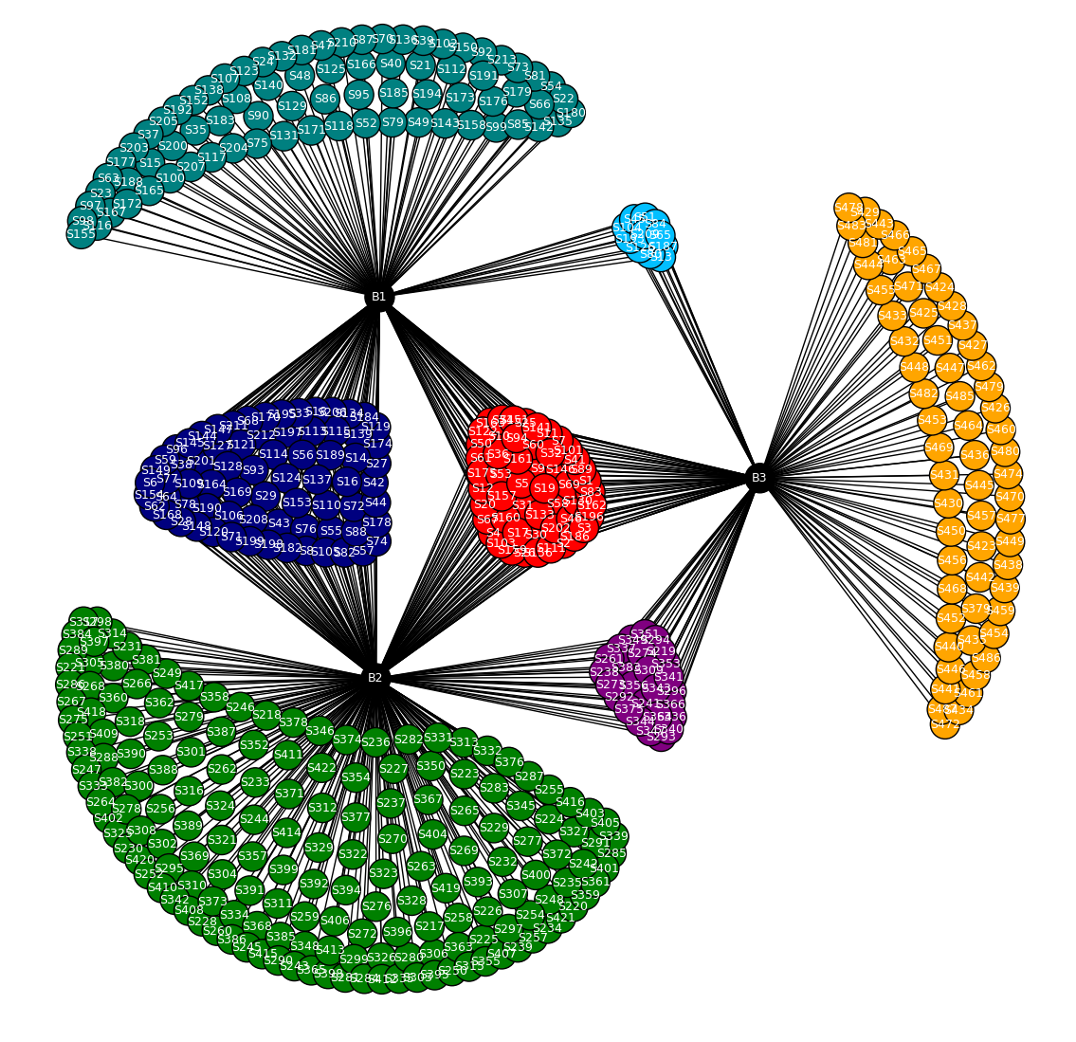}
    \parbox{\linewidth}{\fontsize{8pt}{10pt}\selectfont \textit{Note.} Red suppliers are connected to all buyers (B1, B2, B3), blue suppliers to B1 and B2, cyan suppliers to B1 and B3, purple suppliers to B2 and B3, cyan suppliers only to B1, green suppliers only to B2, and gold suppliers only to B3.}
    \caption{Network-wide relationships between suppliers and buyers in the case study}
    \label{fig1}
\end{figure}

Our case study originates from the maritime engineering sector, where complex engineering assets are produced. The problem setting involves three warehouses, each with a historical dataset that includes previous transactions with their suppliers. 26\% of suppliers are shared across the warehouses. The data encompasses orders delivered from 2015 to 2022, with all buyers located in the United Kingdom. Fig.~\ref{fig1} visualises the network of relationships between suppliers and buyers. 

A total of 77,526 orders have been given to 268 suppliers over this time period, with approximately 62\% of the orders delayed (Table \ref{tab1}). Our aim is to create a causal model for creating effective interventions to prevent delivery delays at the order stage. 
We use the terms intervention and treatment interchangeably as the former is familiar to supply chain practitioners, whereas the latter is a standard term used in CML research. 

{
\begin{singlespace}
\captionsetup{singlelinecheck = false, justification=justified}
\begin{longtable}{@{}lccc@{}}
\caption{Statistical analysis of used datasets} \label{tab1} \\
\toprule
\multicolumn{1}{c}{\textbf{Statistics}} & \textbf{Buyer B1} & \textbf{Buyer B2} & \textbf{Buyer B3} \arraybackslash \\
\midrule
\endfirsthead

\multicolumn{4}{c}%
{{\bfseries Table \thetable\ continued from previous page}} \\
\toprule
\multicolumn{1}{c}{\textbf{Statistics}} & \textbf{Buyer B1} & \textbf{Buyer B2} & \textbf{Buyer B3} \arraybackslash \\
\midrule
\endhead

\multicolumn{4}{r}{{Continued on next page}} \\
\endfoot

\caption*{}
\endlastfoot

Delayed rate (\%) & 56\% & 51\% & 68\% \\
On-Time rate (\%) & 44\% & 49\% & 32\% \\
Maximum of delays (days) & 1669 & 2227 & 1070 \\
Mean of delays (days) & 121.18 & 68.93 & 64.56 \\
Standard deviation of delays (days) & 160.16 & 95.50 & 87.28 \\
\bottomrule
\end{longtable}
\end{singlespace}
}

During the data preprocessing step, missing and duplicated values were removed. Outliers were kept to preserve information. Categorical variables were encoded using the one-hot encoding method. 

To estimate the effectiveness of interventions, information about the following variables is necessary: the intervention of interest, the observed supply chain outcome and problem characteristics (covariates). For example, in delay prediction, one could use MRP records with information about order properties such as supplier selection, ordered products and quantities, requested delivery time (any of which could become the treatment), the deviation from expected arrival times (the observed outcome), and other variables relating to the order (the covariates). 

In our case study, the outcome of interest is ``Delay'', and treatment variables of interest that could be reasonably controlled by the buyers, are the season the order was given in and whether the supplier supplied to multiple warehouses. The latter decision was made due to a hypothesised effect that if a supplier had competing interests products were more likely to be delayed due to prioritisation. This was linked to the company's interest to explore whether to pursue a unique sourcing policy for other orders. 
 
Intervention and outcome variables, and covariates are detailed in Table \ref{tab2}.

{
\begin{singlespace}
\captionsetup{singlelinecheck = false, justification=justified}

\begin{longtable}{@{}p{3cm}p{2cm}p{9cm}@{}}
\caption{Selected features} 
\label{tab2} \\
\toprule
\centering\textbf{Feature Name} & \centering\textbf{Data type} & \centering\textbf{Description} \arraybackslash \\
\midrule
\endfirsthead

\multicolumn{3}{c}%
{{\bfseries Table \thetable\ continued from previous page}} \\
\toprule
\centering\textbf{Feature Name} & \centering\textbf{Data type} & \centering\textbf{Description} \arraybackslash \\
\midrule
\endhead

Supplier & Categorical & Legal name of supplier. \\

Project & Categorical & Name of the project that the ordered product is used for. \\

Part Description & Categorical & Characteristic description of the ordered product. \\

Quantity & Numerical & Quantity of the ordered product. \\

Price & Numerical & Price for ordered product. \\

Delay & Numerical & Duration of delay in days. \\

Season & Categorical & Quarter of the year \\

Multi & Binary & Depicts whether a supplier supplies to another buyer in the network (1 - supplier supplies to more than one buyer, 0 - supplier supply to only one buyer). \\
\bottomrule
\end{longtable}
\end{singlespace}
}

\subsection{Causal Graph Generation}

To make causal quantities identifiable, we need to assume knowledge about the causal relationships, which can be depicted as directed acyclical graphs (DAGs). Based on the causal graph, causal effects, represented by directed edges, can be specified by a set of equations called structural equations. These equations can therefore be seen as a causal interpretation of DAGs, which additionally allow statements about the distribution of variables of interest under what-if scenarios.

Graphs in CML help the practitioner analyse cause and effect relationships \parencite{Pearl2009Causality}. DAGs are used to illustrate causal relations \parencite{pearl1995causal}, where nodes \textit{i} and \textit{j} represent two variables. Then node \textit{i} causes \textit{j} if a directed edge \textit{E[i,j]} from \textit{i} to \textit{j} exists. In a DAG no loops are permitted. Nodes may be connected through a \textit{path} showing steps of causality involved in a given outcome. Thus every edge in a causal graph indicates a direct or indirect causal effect or correlation through other nodes of the graph. 

A DAG can provide an initial starting point for the CML process by ensuring that selected features accurately reflect the underlying causal structure, thereby facilitating a more accurate estimation of causal effects. Note that the resulting variables identified by the DAG help in selecting relevant features that are then input into the CML process.

\begin{figure}[H]
    \centering
    \includegraphics[width=0.5\linewidth]{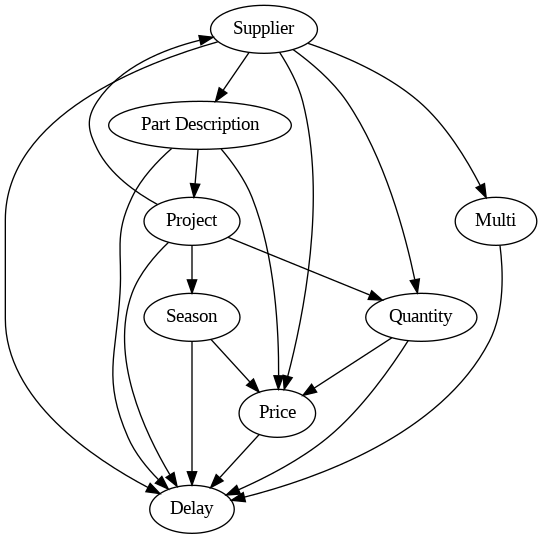}
    \caption{Causal Graph for Delay Prediction based on Practitioner Input}
    \label{figDAG1}
    \caption*{}
\end{figure}

DAGs in CML are usually built using domain knowledge to lay initial assumptions. Following this process in our case study with practitioners from the company resulted in Fig.~\ref{figDAG1}.

Variable ``Multi'' suggests that a supplier's network relationships can significantly influence their operational priorities and capacities, which in turn affects other downstream variables, including ``Delay''. The ``Supplier'' node not only affects ``Part Description'', ``Project'', and ``Quantity'', but also has a direct impact on ``Price'' and ``Delay''. This indicates that changes in the supplier's situation can directly alter the cost of parts and the delivery schedule.

Furthermore, the graph illustrates that ``Part Description'' and ``Project'' also feed into ``Delay'', underscoring how specific parts and their intended use in projects can complicate or expedite delivery timelines. ``Season'' emerges as another critical factor that indirectly impacts ``Delay'' through its influence on ``Quantity'' and ``Price''. This reflects the seasonal variability in supply chain operations, where certain times of the year can affect availability, pricing, and ultimately, delivery schedules.

Even though the DAG in Fig.~\ref{figDAG1} illustrates some of the appropriate causal channels among the variables for use in CML, one might argue that it is subjective. To address this, we experiment with automatic DAG (autoDAG) generation to provide a more objective perspective. Fig. \ref{figaDAGs} shows automatically generated DAGs created by three algorithms: Hill Climbing, Tabu Search, and the Peter Clark (PC) algorithm. We detail their application next.

\subsubsection{Hill Climbing}

Let autoDAG be written as \(G = (V, E)\), where \(E\) is a set of edges and \(V\) is a set of variables defined as \(V = \{X_1, X_2, \ldots, X_n\}\).

The Hill Climbing algorithm is a heuristic search method that can be used to determine the optimal structure of a Bayesian Network from data \parencite{adhitama2022hill}. Note that a Bayesian Network is a type of DAG, designed to visualise Bayesian probability theory. The algorithm's objective function is the K2 score, which measures the compatibility of a graph structure $G$ with the observed data. The aim is to maximise the K2 score, thereby evaluating how well the network structure represents the dependencies in the data. Through a series of local modifications and evaluations, the algorithm converges on a graph that best fits the observed data \parencite{cooper1992bayesian}.

The Hill Climbing algorithm follows these steps:

\begin{enumerate}
    \item Start with an initial graph \(G_0 = (V, E_0)\), where \(E_0\) can be a random set of edges or an empty set of edges.
    
    \item Generate candidate graphs in each iteration, by making local changes to the current graph \(G_t\). The possible changes include:
    \begin{itemize}
        \item Adding an edge: \(G' = (V, E_t \cup \{(X_i \rightarrow X_j)\})\)
        \item Removing an edge: \(G' = (V, E_t \setminus \{(X_i \rightarrow X_j)\})\)
        \item Reversing an edge: \(G' = (V, (E_t \setminus \{(X_i \rightarrow X_j)\}) \cup \{(X_j \rightarrow X_i)\})\)
    \end{itemize}
    
    \item For each candidate graph \(G'\), calculate the K2 score.
    
    \item Among the candidate graphs, select the one that maximises the K2 score. If the highest score for \(G'\) is greater than the current score for \(G_t\), update \(G_t\) to \(G'\).
    
    \item Continue generating candidates, calculating K2 scores, and updating the graph until no further improvements can be made, indicating that a local maximum in the K2 score has been reached.
\end{enumerate}

\begin{figure}[H]
    \centering
    \begin{subfigure}{0.5\textwidth}
        \centering
        \includegraphics[width=\linewidth]{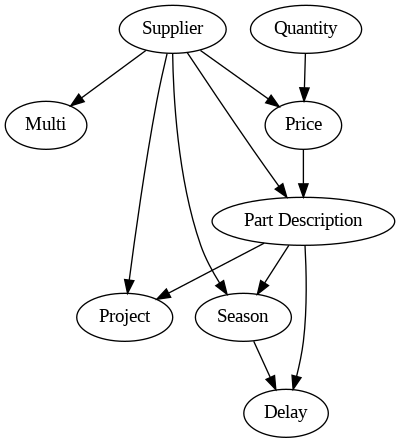}
        \caption{Hill Climbing}
        \label{figDAG2Hill}
    \end{subfigure}\hfill
    \begin{subfigure}{0.5\textwidth}
        \centering
        \includegraphics[width=\linewidth]{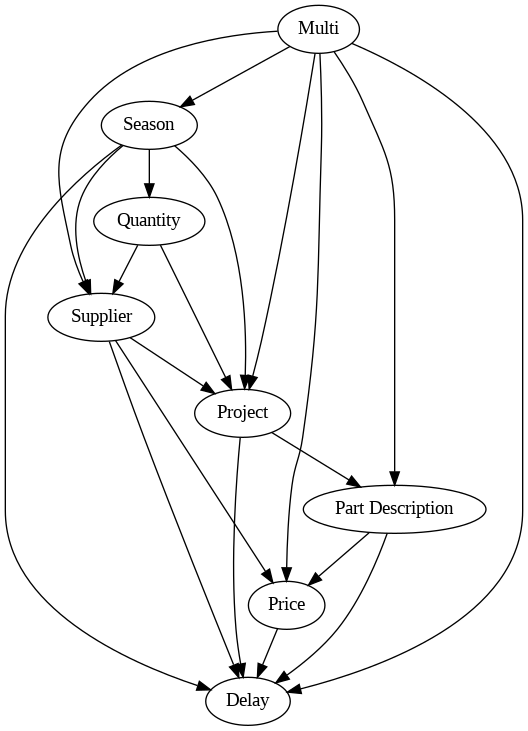}
        \caption{Tabu Search}
        \label{fig:enter-label1}
    \end{subfigure}\hfill
    \begin{subfigure}{0.6\textwidth}
        \centering
        \includegraphics[width=\linewidth]{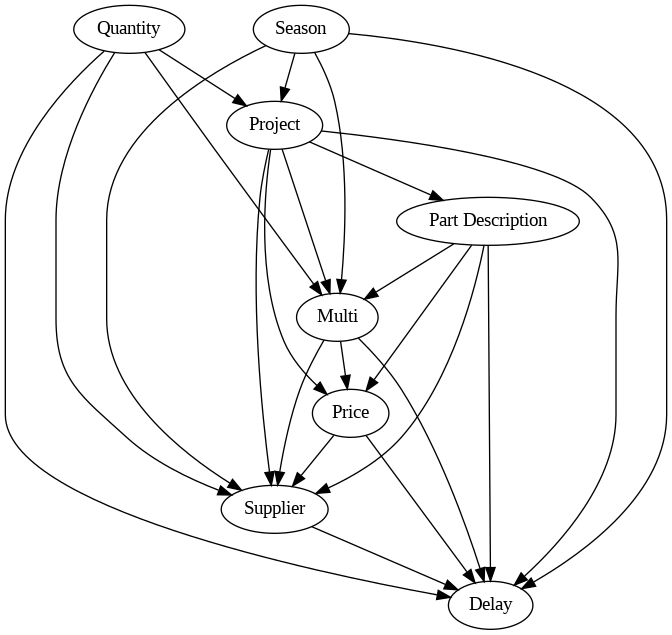}
        \caption{PC algorithm}
        \label{fig:enter-label2}
    \end{subfigure}
    \caption{Automatically Generated Causal Graphs for Delay Prediction}
    \caption*{}
    \label{figaDAGs}
\end{figure}

This approach optimises the search for new structures while exploiting promising ones to avoid overfitting DAGs. From Figure \ref{figaDAGs}a, it can be seen that our algorithm has performed quite well in constructing the domain knowledge DAG. However, it is important to note that the treatment variable ``Multi'' has no direct or indirect effect on any other variables affecting the outcome variable. This is monitored in subsequent DAG generations to decide on keeping the variable in the model. 

\subsubsection{Tabu Search}

The Tabu Search algorithm is a metaheuristic optimisation technique aimed at solving combinatorial optimisation problems \parencite{airoldi2006markov}. Unlike other local search methods, Tabu Search employs a memory structure to avoid revisiting recently explored solutions, thus escaping local optima and enhancing the search process \parencite{kobayashi2021selecting}. In constructing a DAG from observational data, the primary goal is to identify a graph structure that maximises the Bayesian Information Criterion (BIC). BIC balances model fit and complexity by penalising models with excessive parameters to avoid overfitting \parencite{zhou2023data}.

The Tabu Search algorithm follows these steps:

\begin{enumerate}
    \item Start with an initial graph $G_0 = (V, E_0)$, where $E_0$ is an empty set of edges (in our case, the edges from treatment variables to outcome variable were defined). Initialise a Tabu list to track recently explored solutions.
    \item Create a neighbourhood of the current DAG by making local changes to generate candidate DAGs:
    \begin{itemize}
        \item Adding an edge: $G' = (V, E_t \cup \{(X_i \rightarrow X_j)\})$
        \item Removing an edge: $G' = (V, E_t \setminus \{(X_i \rightarrow X_j)\})$
        \item Reversing an edge: $G' = (V, (E_t \setminus \{(X_i \rightarrow X_j)\}) \cup \{(X_j \rightarrow X_i)\})$
    \end{itemize}
    \item Calculate the BIC score for each candidate DAG $G'$.
    \item Among the candidate DAGs $G'$, select the one that maximises BIC score as the new current DAG. Record the move that led to this DAG in the Tabu list to prevent immediate reversal.
    \item Add the recent move to the Tabu list. Remove the oldest entries if necessary to keep the list manageable.
    \item Allow exceptional moves that may otherwise be in the Tabu list if a candidate solution has a BIC score significantly better than the current best solution.
    \item Repeat the process of neighbourhood search, BIC evaluation, candidate selection, and Tabu list updating until a stopping criterion is met. In our case, the stopping criterion is convergence to a solution where no significant improvement is observed.
    \item After the stopping criterion is met, output the DAG with the highest BIC score encountered during the search.
\end{enumerate}

This method, the results of which are shown in Fig.\ref{figaDAGs}b, is known to escape local optimal solutions and to thoroughly traverse through the solution space for capturing major conditional independencies and causal relationships in the data.

\subsubsection{Peter Clark Algorithm}

The PC algorithm was developed by Peter Spirtes and Clark Glymour \parencite{spirtes1991}. It is a general algorithm to generate DAGs based on observational data. The primary objective of the algorithm is to optimise the DAG structure to accurately represent the conditional interdependencies among variables in the data. This involves minimising false positives (incorrectly adding edges) and false negatives (incorrectly removing necessary edges). This process consists of two main stages: 
\begin{enumerate}
    \item Constructing an undirected graph;
    \item Orienting the edges to form a DAG.
\end{enumerate}

In the first step, the algorithm initialises a complete undirected graph $G$ on the set of variables $V$ \parencite{spirtes1993}. A complete graph means each pair of variables in $V$ is initially connected by an edge. The next step involves testing for conditional independence between pairs of variables. For each pair of variables $(X_i, X_j)$, the algorithm checks if there exists a subset of variables $S_{ij} \subseteq V \setminus \{X_i, X_j\}$ such that $X_i$ and $X_j$ are conditionally independent given $S_{ij}$. If such a set $S_{ij}$ is found, the edge $(X_i, X_j)$ is removed from the graph $G$.

To test conditional independence, the algorithm starts with a graph where each variable is connected to every other variable. Using the Fisher \textit{z}-test, it assesses whether the correlation between $X_i$ and $X_j$, controlling for $S_{ij}$, is statistically different from zero. If the test indicates conditional independence, the edge $(X_i, X_j)$ is removed. This step is necessary in order to remove edges that represent indirect dependencies and simplify the graph.

Once the undirected graph $G$ is constructed, the next stage is to orient the edges to convert it into a DAG. This involves:
\begin{enumerate}
    \item detecting \textit{v}-structures: A \textit{v}-structure occurs if there are two edges $(X_i, X_k)$ and $(X_j, X_k)$ such that $X_i$ and $X_j$ are not conditionally independent given any subset of the other variables. These edges are oriented towards the common node $X_k$, forming the structure $X_i \rightarrow X_k \leftarrow X_j$.
    \item propagating edge directions: The algorithm then applies a series of orientation rules to ensure no cycles are formed, maintaining the graph as acyclic. For example:
    \begin{itemize}
        \item If $X_i \rightarrow X_j$ and $X_j - X_k$, then $X_j \rightarrow X_k$ to prevent the cycle $X_i \rightarrow X_j \rightarrow X_k$.
        \item If $X_i - X_j$ and there is a path $X_i \rightarrow X_k \rightarrow X_j$, then $X_i \rightarrow X_j$.
    \end{itemize}
\end{enumerate}

These rules are applied iteratively until all edges are correctly oriented without forming any cycles. By following these steps, the PC algorithm constructs a DAG that captures the causal structure underpinning the observed data, providing valuable insights into the relationships between variables.

Observing autoDAGs created by the PC algorithm and Tabu Search we see a reflection of the cause and effect relationships between variables in our case study. Despite some differences between the graphs, they can still be rationally explained. For instance, the Tabu Search graph shows a direct influence of ``Season'' on ``Quantity''. This is logical because seasonality can significantly affect the quantity of products ordered (e.g, product availability). On the other hand, the PC algorithm graph does not show this direct influence. Instead, we see that ``Quantity'' affects ``Project''. In real world scenarios, this is plausible because the quantity of parts needed for a project is often predefined. Therefore, seasonality may not directly impact the quantity ordered for a project, as the requirements are already established based on project specifications and timelines.

This example of differences shows that despite the different algorithmic approaches, both graphs can be rationally explained based on domain knowledge. The autoDAGs confirm the appropriate causal channels among the variables for use in CML.

\subsection{Development of the Interactive Regression Model}

As mentioned earlier, CML involves the application of a suitable ML model through a DML framework to estimate the relationship between the outcome variable and treatment as well as covariates. In our case study, decisions about mitigating delivery risk are based on binary characteristics \( X \), such as supplier selection. We thus choose an Interactive Regression Model (IRM) that can handle such a form of treatment. IRM takes the following form \parencite{chernozhukov2018double}:

\begin{equation}
\begin{aligned}
Y & = g_0(D, X) + U, \quad \mathbb{E}[U \mid X, D] = 0, \\
D & = m_0(X) + V, \quad \mathbb{E}[V \mid X] = 0.
\end{aligned}
\end{equation} 

\noindent where $Y$ is the outcome variable and $D$ is the binary and heterogeneous treatment. $X$ represents a high-dimensional vector of confounding covariates $(X_1, \ldots, X_p)$. $U$ and $V$ signify stochastic errors \parencite{Bach_2022_JMLR}. Conditional expectations \parencite{chernozhukov2018double} are as follows:

\begin{equation}
\begin{aligned}
g_0(D, X) & =\mathbb{E}[Y \mid X, D], \\
m_0(X) & =\mathbb{E}[D \mid X]=P(D=1 \mid X)
\end{aligned}
\end{equation}

\noindent These are unknown and might be complex functions of \( X \). In this model, the target parameter of interest are the average treatment effect, known as ATE \parencite{schacht2023causally}:

\begin{equation}
\theta_0=\mathbb{E}\left[g_0(1, X)-g_0(0, X)\right]
\end{equation}

ATE conceptualises potential outcomes, which are the outcomes that would hypothetically be observed if a certain intervention was applied. ATE measures effects at the level of the study sample. By comparing the average outcome for cases receiving the intervention versus those which do not (control group), ATE helps in understanding how effective a treatment is, on average, across a specific sample grouping. In other words, ATE measures the mean difference in outcomes between units assigned to the treatment and those assigned to the control across the entire population \parencite{pirracchio2016propensity}.

In the case of IRM, the score for an ATE estimator is given by the linear formula \parencite{schacht2023causally}:

\begin{equation}
\label{eq:ate}
\begin{aligned}
\psi\left(W_i ; \theta, \eta\right):= & \psi_a\left(W_i, \eta\right) \theta+\psi_b\left(W_i, \eta\right) \\
= & -\theta+g\left(1, X_i\right)-g\left(0, X_i\right) \\
& +\frac{A_i\left(Y_i-g\left(1, X_i\right)\right)}{m\left(X_i\right)}-\frac{\left(1-A_i\right)\left(Y_i-g\left(0, X_i\right)\right)}{1-m\left(X_i\right)},
\end{aligned}
\end{equation}

For a more granular view, we can estimate the conditional average treatment effect (CATE), which is the effect of an intervention for a particular subgroup of samples defined by the covariates.  Understanding the heterogeneity in intervention effects can inform about subgroups of cases where interventions are not effective, and is relevant for individualizing interventions for specific cases. CATEs are the average effect of a treatment on an outcome for a specific subgroup within the population \parencite{jacob2021cate}. The IRM model also offers the ability to calculate CATE as: 

\begin{equation}
\theta_0(x)=\mathbb{E}[Y(1)-Y(0) \mid X=x]
\end{equation}

\noindent where \(\tilde{X}\) represents a set of covariates that are not necessarily included in \( X \) \parencite{semenova2021}. 

Finally, the IRM model can be used to estimate deterministic binary treatment policies using classification trees \parencite{bach2024doubleml}. Note that a deterministic binary treatment policy refers to a decision-making rule that assigns one of the two possible treatments based on observable characteristics. The policy estimation process is informed by the components \(\psi_b\left(W_i, \eta\right)\) in equation~\ref{eq:ate} \parencite{athey2021policy}. \textcite{athey2021policy} proposed to estimate the treatment assignment rule as:

\begin{equation}
\hat{\pi} = \arg \max_{\pi \in \Pi} \sum_{i=1}^{n} (2\pi(X_i) - 1) \psi_b(W_i, \hat{\eta})
\end{equation}

where the weights \( \lambda_i \) are defined as \( |\psi_b(W_i, \hat{\eta})| \) and the target \( H_i \) is the sign of \( \psi_b(W_i, \hat{\eta}) \). This method aligns with creating a decision framework that emphasises the weight and direction indicated by the treatment effect estimations \( \psi_b \) \parencite{athey2021policy}.

\section{Experimental evaluation}

\subsection{Experimental setup}

Experiments were conducted using Python on an Apple Macbook M1 with 8GB RAM.  \texttt{XGBoost} was employed as the main classifier and regressor for training the IRM model. By employing XGBRegressor and XGBClassifier, the relationships between variables can be estimated in a highly flexible manner, avoiding the imposition of rigid assumptions on the functional forms of \( g \) and \( m \). This flexibility is crucial for capturing complex patterns and interactions within the data, significantly enhancing the reliability and accuracy of the estimation process. 

To identify the optimal hyperparameters, we utilised GridSearchCV within scikit-learn version 1.4.2~\parencite{scikit-learn} , which systematically iterates through multiple combinations of parameter settings, performing cross-validation to determine the best performance configuration. The default 5-fold cross-validation was used.

We used the package \texttt{DoubleML} which is built on top scikit-learn and implements DML and provides estimation of causal effects in different models e.g., the partially linear regression model (PRL), the partially linear instrumental variable regression model (PLIV), the interactive regression model (IRM), and the interactive instrumental variable regression model (IIVM).

\subsection{Results}

ATE estimates are presented in Table \ref{tab3}. As ATE provides population level patterns, we can interpret the result as follows. When a buyer is purchasing from a supplier that is connected to another warehouse, the duration of the delay will increase on average by approximately 17 days. This result is statistically significant based on p-values and t-statistics. Hence, the original hypothesis on multiple warehouses being served by the supplier contributing to a delay cause can be considered correct. 
Note that even if we have actually noted this as causality, the reasons behind this still need to be determined. \textit{Interpreting} results through causal links does not always equate to fully \textit{explaining} the reasoning and causal path behind an outcome, since interpretability and explainability do not necessarily overlap in the context of AI~\parencite{Antoniou2022A}. As appreciated by the practitioner, the effect might indeed be due to an inherent prioritisation on the side of the supplier, but it may also be due to the supplier batching production, e.g. due to set up and delivery costs involved, or even as supplier has exceeded capacity due to multiple requests. The actual root causes of supplier behaviour and potential solutions need to be explained through discussions with the supplier. 

{\begin{singlespace}
\captionsetup{singlelinecheck = false, justification=justified}

\begin{longtable}{@{}>{}p{3cm}>{\centering\arraybackslash}p{3cm}>{\centering\arraybackslash}p{3cm}>{\centering\arraybackslash}p{3cm}>{\centering\arraybackslash}p{3cm}@{}}
\caption{Average Treatment Effects} 
\label{tab3} \\
\toprule
\textbf{Treatment} & \textbf{Coef} & \textbf{t-statistics} & \textbf{P-value} & \textbf{Std error} \\
\midrule
\endfirsthead

\multicolumn{5}{c}%
{{\bfseries Table \thetable\ continued from previous page}} \\
\toprule
\textbf{Treatment} & \textbf{Coef} & \textbf{t-statistics} & \textbf{P-value} & \textbf{Std error} \\
\midrule
\endhead

\multicolumn{5}{r}{{Continued on next page}} \\
\endfoot
\caption*{}
\endlastfoot

Multi & 16.74 & 17.59 & 2.96 $\times$ 10$^{-69}$ & 0.534 \\
First quarter & -5.90 & -9.72 & 2.38 $\times$ 10$^{-22}$ & 0.662 \\
Second quarter & 8.22 & 13.13 & 2.14 $\times$ 10$^{-39}$ & 0.607 \\
Third quarter & 10.12 & 16.20 & 4.71 $\times$ 10$^{-59}$ & 0.624 \\
Fourth quarter & -18.70 & -27.97 & 3.49 $\times$ 10$^{-172}$ & 0.669 \\
\bottomrule
\end{longtable}
\end{singlespace}
}

Another observation is that buying parts in the second and third quarter will increase the delay by 8 and 10 days, respectively. In contrast, buying products during the first and fourth quarters will decrease the delay by approximately 6 and 19 days, respectively. These seasonal variations are quite significant. Seasonal order placement therefore needs to play an important role in scheduling deliveries from the supplier which may in turn impact the buyers' own production and delivery schedules. 

These results based on ATE and supported by the DAG analysis, provide a clear understanding of the causal factors affecting delivery delays. They underscore the importance of considering both the supplier's network and seasonal timing when planning orders, thus enabling more strategic decision-making to mitigate risks in the supply chain.

\begin{figure}[H]
    \centering
    \begin{subfigure}[b]{0.45\linewidth}
        \centering
        \includegraphics[width=\linewidth]{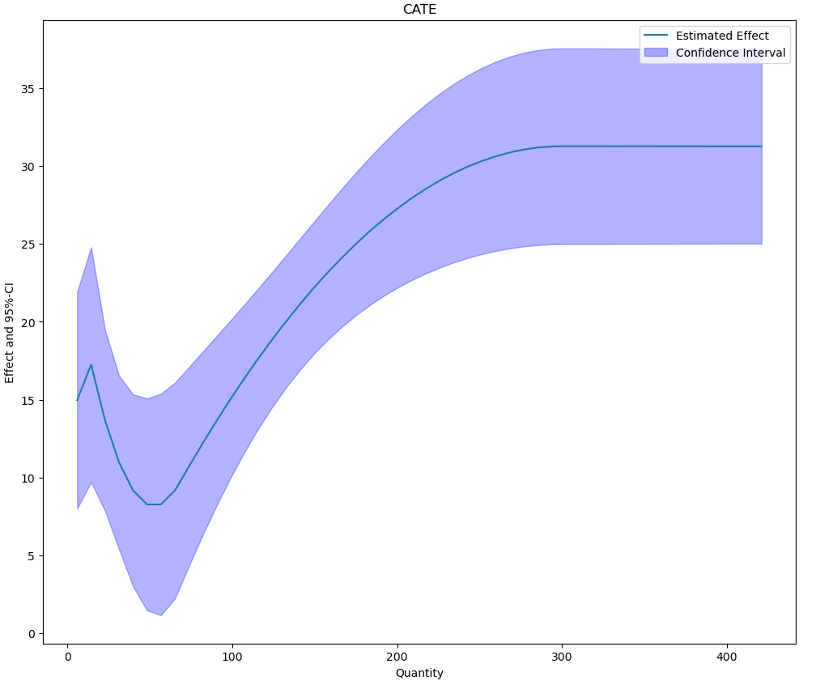}
        \caption{Multi}
        \label{fig:ct_link}
    \end{subfigure}
    \hfill
    \begin{subfigure}[b]{0.45\linewidth}
        \centering
        \includegraphics[width=\linewidth]{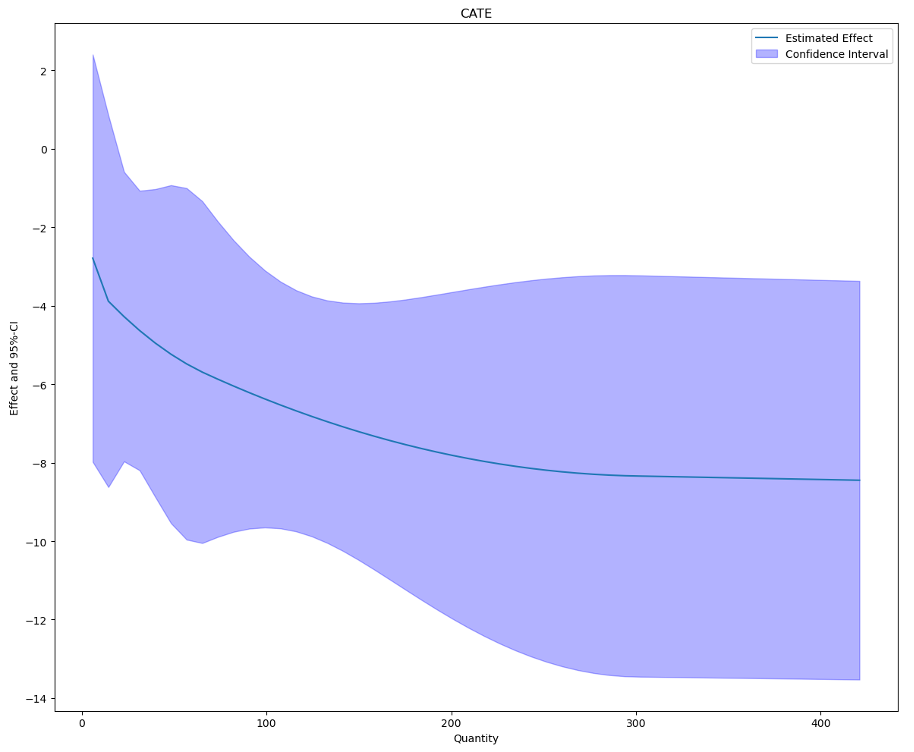}
        \caption{First quarter}
        \label{fig:ct_q1}
    \end{subfigure}
    \vskip\baselineskip
    \begin{subfigure}[b]{0.45\linewidth}
        \centering
        \includegraphics[width=\linewidth]{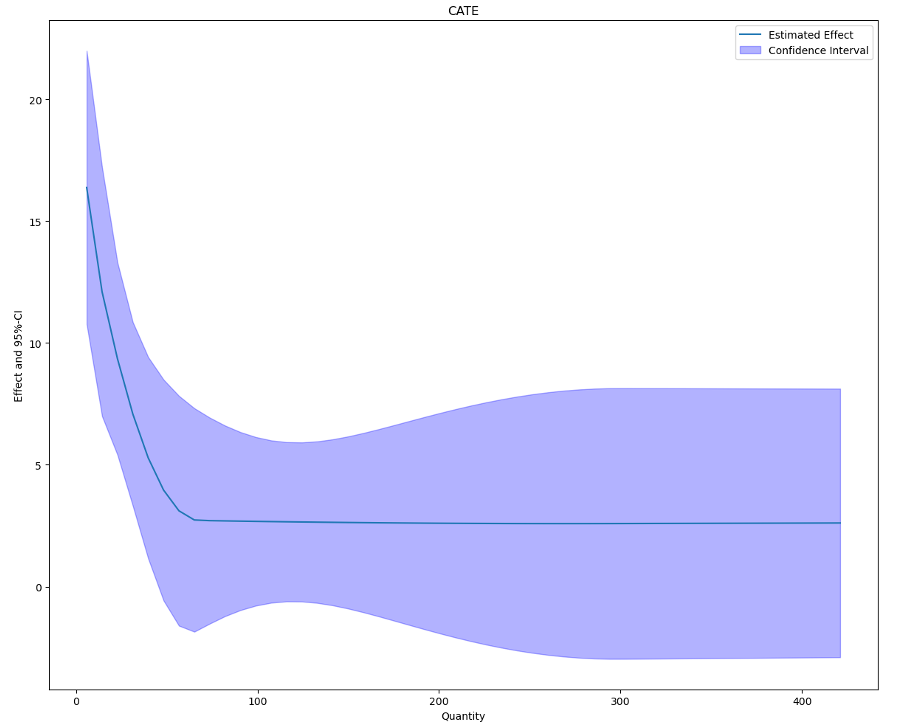}
        \caption{Second quarter}
        \label{fig:ct_q2}
    \end{subfigure}
    \hfill
    \begin{subfigure}[b]{0.45\linewidth}
        \centering
        \includegraphics[width=\linewidth]{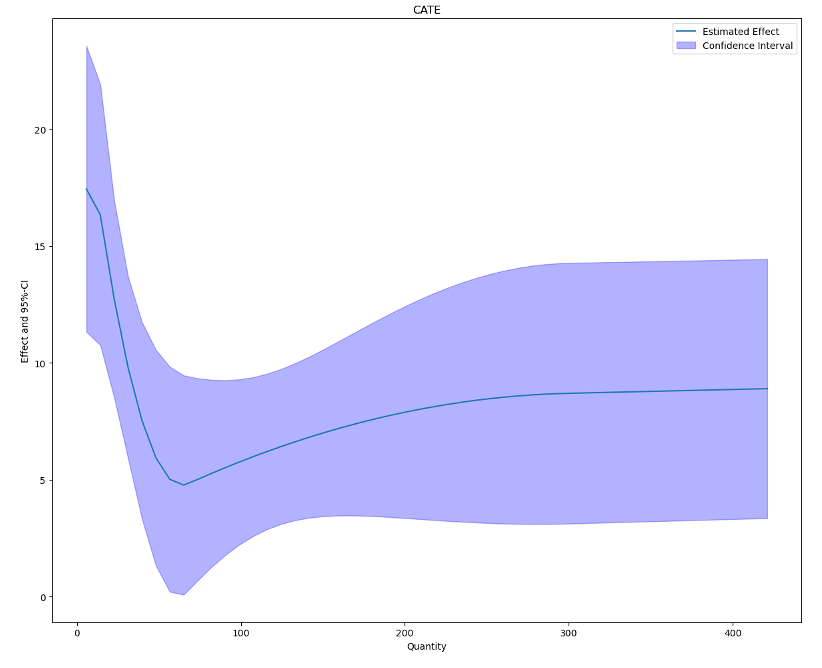}
        \caption{Third quarter}
        \label{fig:ct_q3}
    \end{subfigure}
    \vskip\baselineskip
    \begin{subfigure}[b]{0.45\linewidth}
        \centering
        \includegraphics[width=\linewidth]{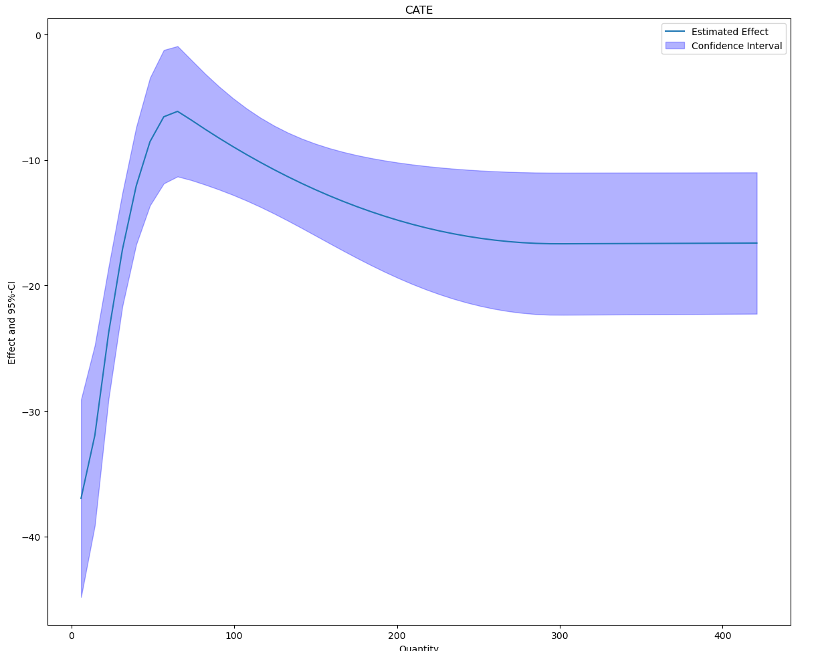}
        \caption{Fourth quarter}
        \label{fig:ct_q4}
    \end{subfigure}
    \caption{Conditional Average Treatment Effects}
    \label{figCATEs}
\end{figure}

An analysis using CATE provides a more granular interpretation based on order specifics. Fig. \ref{figCATEs} represents one-dimensional CATEs that depend on the covariate ``Quantity''. To estimate the effect, we used a B-splines basis with 5 degrees of freedom. Fig. \ref{figCATEs}a shows that for small quantities, the effect is uncertain and highly variable. For quantities up to 100-200, the effect increases. For quantities greater than 200, the effect becomes constant, though there is more uncertainty for higher quantities. In the first, second, and third quarters, we can see that the effect decreases and then becomes constant, with increased uncertainty. In contrast, the effect in the fourth quarter increases and then also becomes constant, but there is less uncertainty compared to other quarters.

By estimating the CATEs, we gain insights into how specific covariates, such as the order quantity, influence the effect of interventions across different scenarios. This complements the ATE analysis by revealing heterogeneity in treatment effects and highlighting the importance of tailoring interventions to specific contexts. Through this approach, supply chain practitioners can better understand the nuances of how different factors interact and affect outcomes, thereby improving the precision and effectiveness of their decision-making processes.

\begin{figure}[H]
    \centering
    \begin{subfigure}[b]{0.45\linewidth}
        \centering
        \includegraphics[width=\linewidth]{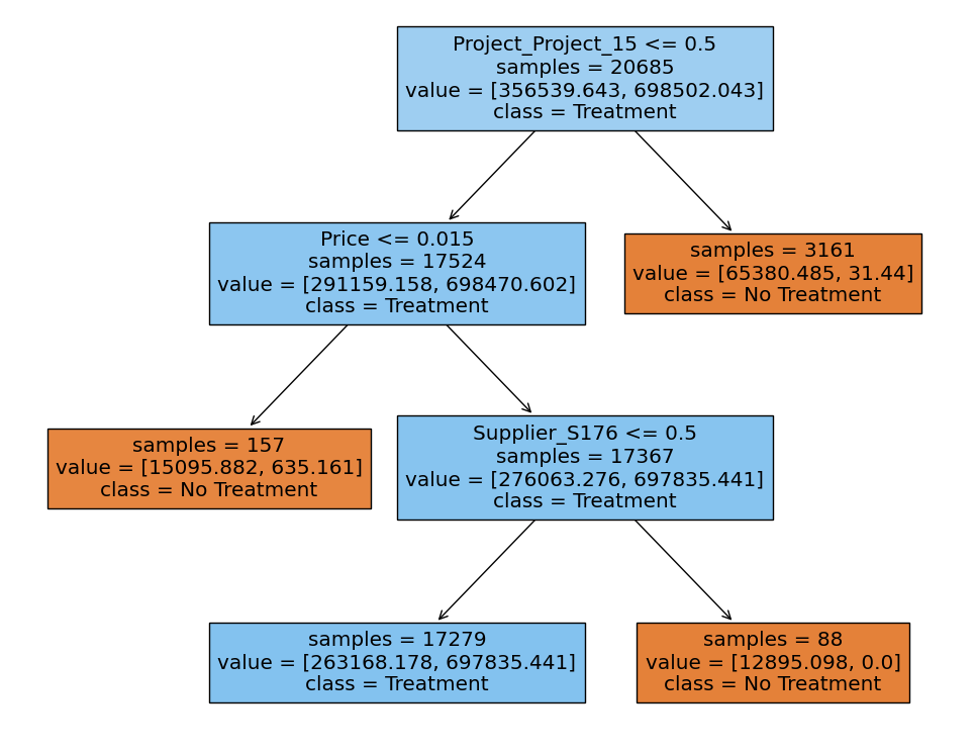}
        \caption{Multi}
        \label{fig:pt_link}
    \end{subfigure}
    \hfill
    \begin{subfigure}[b]{0.45\linewidth}
        \centering
        \includegraphics[width=\linewidth]{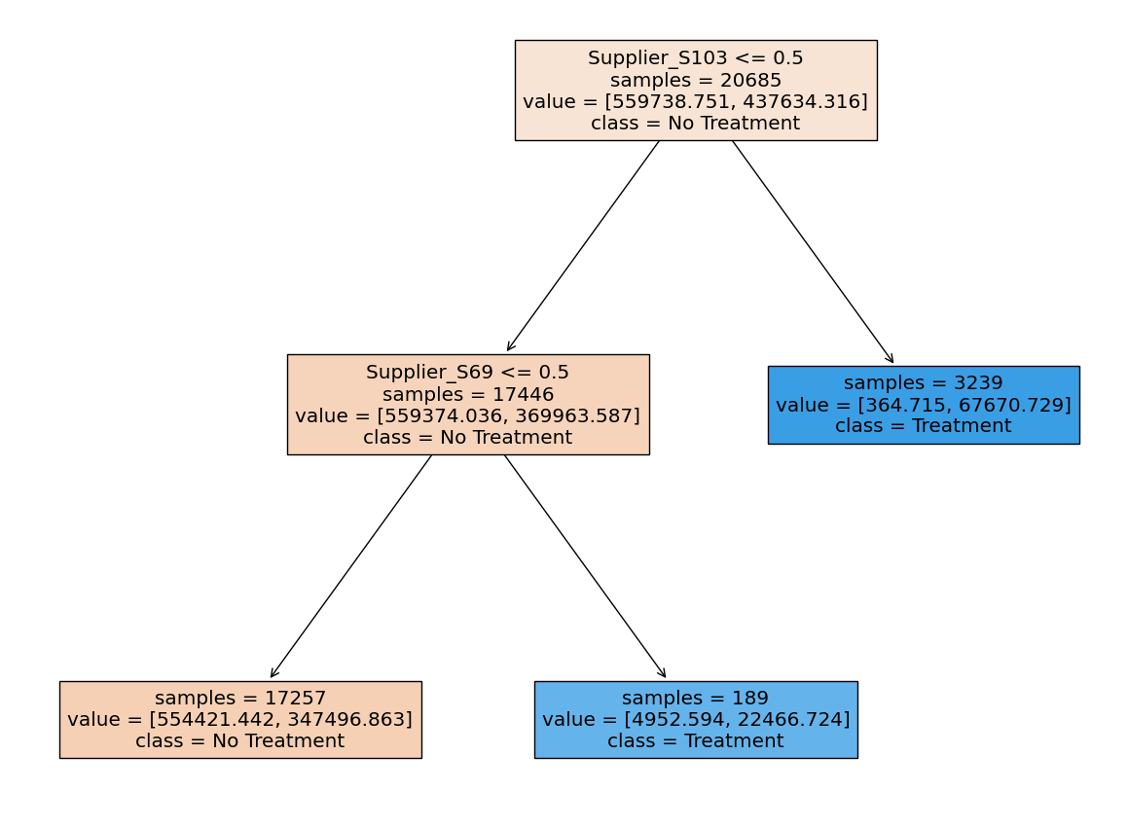}
        \caption{First quarter}
        \label{fig:pt_q1}
    \end{subfigure}
    \vskip\baselineskip
    \begin{subfigure}[b]{0.45\linewidth}
        \centering
        \includegraphics[width=\linewidth]{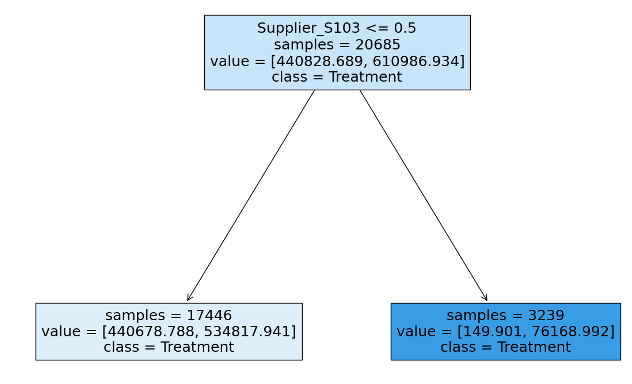}
        \caption{Second quarter}
        \label{fig:pt_q2}
    \end{subfigure}
    \hfill
    \begin{subfigure}[b]{0.45\linewidth}
        \centering
        \includegraphics[width=\linewidth]{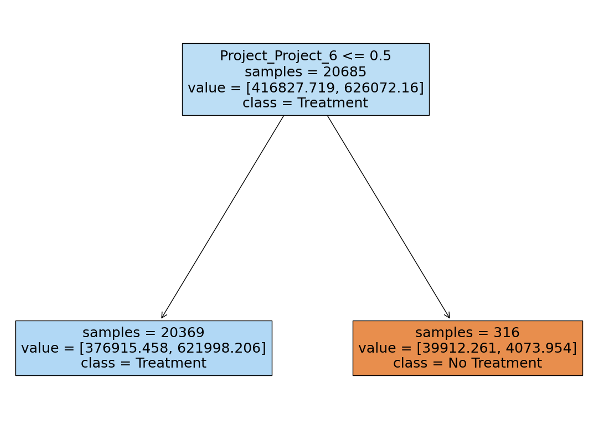}
        \caption{Third quarter}
        \label{fig:pt_q3}
    \end{subfigure}
    \vskip\baselineskip
    \begin{subfigure}[b]{0.45\linewidth}
        \centering
        \includegraphics[width=\linewidth]{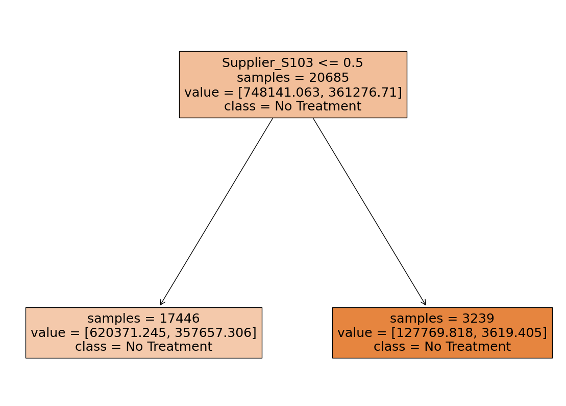}
        \caption{Fourth quarter}
        \label{fig:pt_q4}
    \end{subfigure}
    \caption{Policy Trees}
    \label{figPolicy}
\end{figure}

Next, we estimated deterministic binary treatment policies. Fig. \ref{figPolicy} represents policy trees for treatment variables. Starting with the ``Multi'' variables, we observe in Fig. \ref{figPolicy}a that the first split is based on the status of Project 15. Then, if a practitioner is ordering for Project 15 and the price of the order is less than or equal to 0.015 (normalised), we should use suppliers connected to only one warehouse for this part. In contrast, when the price is greater than 0.015, we could consider suppliers that serve multiple warehouses. However, if supplier S176 is used, deliveries should be supplied by suppliers connected to only one warehouse. 

Fig. \ref{figPolicy}b shows the policy tree for the treatment ``First quarter'', with splits based on the usage of suppliers S103 and S69. In the case of the second, third, and fourth quarters, there is only one split based on the status of Project 6 and Supplier S103. Note that in the case of the ``Fourth quarter'', the policy tree suggests that Supplier S103 should not be used during this time, whereas, during the second quarter of the year, this particular supplier should be used exclusively.

\section{Conclusions \& managerial implications}

Causal machine learning in supply chain management represents a significant advancement in tackling the complexities and challenges of modern businesses. Unlike traditional ML models that primarily focus on correlations with the primary aim of maximising predictive accuracy, this approach delves into the causal relationship between control variables and a change in an outcome. Through this process, CML allows policies to be drawn from data, in order to influence a predicted outcome variable. 

In this paper we illustrated the application of CML in the context of supply chain risk management through an empirical case study in the maritime engineering sector, to predict and mitigate supply chain risk, in relation to delivery delays. 

Our results indicate the significant potential of using CML to generate effective interventions for minimising delays. For example, we found that buyers who utilise suppliers that serve multiple warehouses experience longer delivery delays compared to those who use suppliers dedicated to a single buyer and that the season in which an order is placed strongly impacts upon its potential delay. Such insights then led to designing specific policies for suppliers, using control variables to mitigate risks. 

The use of DAGs played a crucial role in our study as a means to explicitly formulate causal problems. DAGs define causal relationships between variables and warranty that our approach is correctly structured in understanding how different factors influence supply chain delays. The graphical representation through DAGs also assists in identifying and focusing on the most relevant causal pathways for maximal intervention impact. To complement the subjective insights gained from practitioner knowledge, we also employed automated DAG generation methods such as Hill Climbing, Tabu Search and PC algorithm. These balance out the subjective perspective of expert knowledge by removing bias and acting as a means to validate initial assumptions. This dual approach enhances the strength and robustness of causal inferences and interventions.

Implementing CML for understanding delivery delays has important managerial implications as it can lead to improved decision-making, enhanced efficiency, and reduced costs. By identifying and leveraging causal relationships, managers can develop more targeted and effective strategies to address specific issues within the supply chain. CML should be used in tandem with traditional ML whereby traditional ML allows for accurate forecasting whereas CML allows for effective interventions by changing causal parameters. This complementary approach ensures that businesses can not only predict potential disruptions but also also apply targeted interventions to mitigate their impact, thereby enhancing the overall performance of supply chain, and thus its competitive advantage. 

Our study has limitations that open up avenues for further research. Primarily, the case-study nature of our research limits the applicability of our findings to other sectors, suggesting a need for additional research to validate the effectiveness of CML across different industrial contexts. Applying CML to other SCRM-related issues, such as inventory management, predictive maintenance and quality control would be particularly noteworthy as these areas provide ample space for interventions given the availability of rich datasets. 

Causal Federated Learning could provide another promising avenue for future research, where actions of multiple self-interested but interdependent agents (firms) in a supply chain can be considered within a CML framework to help identify supply chain failures in a privacy preserving manner. This method could facilitate collaborative yet secure data analysis across different entities, potentially uncovering systemic inefficiencies and opportunities for interventions.

Concluding, the adoption of causal machine learning in supply chain risk management holds the potential to revolutionise the field by providing deeper insights and more precise interventions. As businesses continue to face increasingly complex issues affecting supply chains, the integration of CML can drive significant advancements in efficiency, resilience, durability, and overall performance. Further exploration and validation of CML across diverse industrial contexts will be critical to unlocking its full potential and fostering innovation in supply chain risk management.     

\section*{Acknowledgments}

This research has been supported by funds granted by the Minister of Science of the Republic of Poland under the ``Regional Initiative for Excellence'' Programme for the implementation of the project “The Poznań University of Economics and Business for Economy 5.0: Regional Initiative – Global Effects (RIGE)”

\printbibliography[title={References}, heading=bibintoc]

\end{document}